\documentclass[11pt,a4paper]{article}
\usepackage[hyperref]{acl2019}
\usepackage{times}
\usepackage{latexsym}

\usepackage{url}

\usepackage{xcolor}
\usepackage{todonotes}
\usepackage{amssymb}

\aclfinalcopy % Uncomment this line for the final submission

\newcommand{\fatterdot}{\raisebox{0.25ex}{\tikz\filldraw[black,x=2pt,y=2pt] (0,0) circle (1);}}

\title{\textsc{Same Side Stance Classification Task}: Facilitating Argument Stance Classification by Fine-tuning a BERT Model}

\author{
  Stefan Ollinger, Lorik Dumani, Premtim Sahitaj, Ralph Bergmann, Ralf Schenkel \\
  University of Trier\\
  D-54286 Trier \\
  \texttt{stefan.ollinger@gmx.de}\\
  \texttt{\{dumani, s4prsahi, bergmann, schenkel\}@uni-trier.de}
}
\date{}

\begin{document}
\maketitle
\begin{abstract}
Research on computational argumentation is currently being intensively investigated. The goal of this community is to find the best pro and con arguments for a user given topic either to form an opinion for oneself, or to persuade others to adopt a certain standpoint.
While existing argument mining methods can find appropriate arguments for a topic, a correct classification into pro and con is not yet reliable.
The same side stance classification task provides a dataset of argument pairs classified by whether or not both arguments share the same stance
and does not need to distinguish between topic-specific pro and con vocabulary but only the argument similarity within a stance needs to be assessed.
The results of our contribution to the task are build on a setup based on the \textsc{BERT} architecture. We fine-tuned a pre-trained \textsc{BERT} model for three epochs and used the first 512 tokens of each argument to predict if two arguments share the same stance.
%This paper describes the setup and results of our contribution to the same side (stance) classification task. In our implementation we fine-tuned \textsc{BERT} for three epochs and used the first 512 tokens to decide whether two given arguments have the same stance. We evaluated our methods within topics and across topics.
\end{abstract}

\section{Introduction}

Argumentation is an activity in everyday human life. We argue in domains such as health, law and politics either trying to find standpoints which are acceptable by being supported by reasons or to persuade others to a certain point of view and if necessary to carry out certain actions.
Computational Argumentation (CA) aims to find argument representations and models which are well suited to do computation with arguments. CA is a new and fast growing field of research.
In the simplest case, an \textit{argument} is a \textit{claim} supported or opposed by at least one \textit{premise}~\cite{DBLP:journals/ijcini/PeldszusS13}.
An example of a claim $c$ could be ``\textit{We need to abolish nuclear energy}'', examples of premises that support and oppose this claim could be $p_1 =$ ``\textit{Renewable energy sources will eventually be able to replace fossil fuel and nuclear energy}'' and $p_2 =$ ``\textit{Nuclear energy is a cheap alternative to fossil fuels}'', respectively.
Common tasks in CA include argument mining (AM) and argument retrieval (AR). AM reconstructs arguments from textual sources, e.g. in form of an argument graph. AR finds all relevant arguments for a topic.
Existing argument search engines like \textsc{Args}\footnote{\url{www.args.me}} or \textsc{ArgumenText}\footnote{\url{www.argumentext.de}} search for the best supporting and opposing premises for a user query on a usually controversial topic and list them separately in pro and con.
The correct classification of stances is therefore a fundamental task in computational argumentation.
However, a short-coming of current stance classification algorithms is that their classifiers must be trained for a particular topic, i.e. they cannot be reliably applied across topics~\cite{SameSideWebisTask}.

% Observe that same side classification needs not to distinguish between topic-specific pro- and con-vocabulary — ''only'' the argument similarity within a stance needs to be assessed. I.e., same side classification can probably be solved independently of a topic or a domain, so to speak, in a topic-agnostic fashion. We believe that the development of such a technology has game-changing potential. It could be used to tackle, for instance, the following tasks: measure the bias strength within an argumentation, structure a discussion, find out who or what is challenging in a discussion, or filter wrongly labeled arguments in a large argument corpus—without relying on knowledge of a topic or a domain.

In the task $\gg$\textsc{same side (stance) classification}$\ll$ a simplified variant is to be examined, namely whether two given arguments to a topic have the same stance.
For example, $p_1$ and $p_2$ have different stances, but $p_1$ and $p_3 =$ ``\textit{The danger from radioactive contamination should be avoided}'' would have the same stance to the topic \textit{nuclear energy}.
The particular difficulty lies in the fact that $p_1$ and $p_3$ are syntactically very different. So we have to decide on a semantic level whether the stances are the same.

In this paper we present a method where we fine-tune a pre-trained \textsc{BERT}~\cite{DBLP:conf/naacl/DevlinCLT19} model to decide whether two arguments have the same stance.
%For within-topics we achieved an accuracy of \textcolor{red}{X} and an F1 score of \textcolor{red}{X}, and for cross-topics we achieved an accuracy of \textcolor{red}{X} and an F1 score of \textcolor{red}{X}.
In Section~\ref{relatedWork} we discuss related work. In Section~\ref{dataset} we specify the dataset. Then, in Section~\ref{evaluation} we describe the implementation and the evaluation of our approach. Finally, Section~\ref{conclusion} concludes the paper.

\section{Related Work}
\label{relatedWork}

% BERT
In our implementation we make use of \textsc{BERT}~\cite{DBLP:conf/naacl/DevlinCLT19} which achieved state-of-the-art results in many NLP tasks such as Natural Language Inference (MNLI), semantic textual similarity (STS), and others.
\textsc{BERT} makes use of the Transformer architecture~\cite{DBLP:conf/nips/VaswaniSPUJGKP17}, more precisely it applies a bidirectional masked language model training to the architecture. Contrary to previous embedding techniques like \textsc{Word2Vec}~\cite{DBLP:conf/nips/MikolovSCCD13} its mechanism learns contextualized representations of words in a text.

% UKP
\cite{DBLP:conf/emnlp/StabG14} address argumentative relation classification. The relation between two argument components is divided in \textit{support} and \textit{non-support} classes. Therefore a range of structural, lexical and syntactic features are defined and extracted for an argument component pair. The classification is done with a SVM.

% IBM
\citet{DBLP:conf/eacl/SlonimBSBD17} address stance classification of premises towards a claim topic. Here, the classification task is divided further into simpler sub-tasks.
\citet{DBLP:conf/argmining/Bar-HaimEJS17} extend this work by a more extensive sentiment lexicon and contextual features.

Most presented approaches classify the relation of a premise towards a claim. In contrast to the same stance classification the relation between two premises is considered. Further we do not apply feature engineering, but rely on the neural network to extract good features.

\section{The provided Dataset}
\label{dataset}
The arguments from the provided dataset were extracted from the four web sources \url{idebate.org}, \url{debatepedia.org}, \url{debatewise.org} and \url{debate.org}.
Each instance consists of the seven fields that are depicted in Table~\ref{tab:instancesInDataset}.

\begin{table}[htb]
    \centering
    \begin{tabular}{|p{0.3\linewidth}|p{0.6\linewidth}|}
        \hline
        \textbf{Label}              &   \textbf{Description}\\
        \hline
        \textit{id}                 &   The id of the instance\\
        \hline
        \textit{topic}              &   The title of the debate. It can be a general topic (e.g. abortion) or a topic with a stance (e.g. abortion should be legalized).\\
        \hline
        \textit{argument1}          &   A pro or con argument related to the topic.\\
        \hline
        \textit{argument1\_id}      &   The ID of argument1.\\
        \hline
        \textit{argument2}          &   A pro or con argument related to the topic.\\
        \hline
        \textit{argument2\_id}      &   The ID of argument2.\\
        \hline
        \textit{is\_same\_stance}   &   True or False. True in case argument1 and argument2 have the same stance towards the topic and False otherwise.\\
        \hline
    \end{tabular}
    \caption{Fields of each instance in the dataset. Source:~\cite{SameSideWebisTask}}
    \label{tab:instancesInDataset}
\end{table}

The two most discussed topics ``abortion'' and ``gay marriage'' were chosen and two experiments were set-up for the same side stance classification task.
The first experiment addresses the classification within topics and consists of a training set with arguments for a set of topics (abortion and gay marriage) and a test set with arguments related to the same set of topics.
Table~\ref{tab:overviewOfTheDataWithinTopics} illustrates an overview of the data within topics.

\begin{table}[htb]
    \centering
    \begin{tabular}{|l|c|c|}
        \hline
                                &   \textbf{topic:}     &   \textbf{topic:}\\
        \textbf{class}          &   \textbf{abortion}   &   \textbf{gay marriage}\\
        \hline
        \textit{Same Side} 	    &   20,834              &   13,277\\
        \textit{Different Side} &   20,006              &   9,786\\
        \textit{Total}          &   40,840              &   23,063\\
        \hline
    \end{tabular}
    \caption{Overview of the data within topics. Source:~\cite{SameSideWebisTask}}
    \label{tab:overviewOfTheDataWithinTopics}
\end{table}

For the second experiment, which addresses the classification across topics, the training set contains arguments for a topic (abortion) and the test set arguments are related to another set of topics.
The class \textit{Same Side} contains 31,195 instances, the class \textit{Different Side} 29,853.

\section{Evaluation}
\label{evaluation}

In this section we describe the experimental setup and evaluate our approach utilizing \textsc{BERT} and compare it to a \textsc{SVM} baseline.

\subsection{Hypotheses}

In order to measure the performance of our approach, the following hypotheses were formulated and are subject of this evaluation:

\begin{itemize}
\item \textbf{H1:} A Transformer-based sequence classification improves upon the SVM baseline.
\item \textbf{H2:} The \textit{large} Transformer model outperforms the smaller \textit{base} model.
\item \textbf{H3:} Longer input sequences yield better classification than shorter sequence lengths.
\item \textbf{H4:} Classification of full sentences performs better than including partial input sentences.
\end{itemize}

\subsection{Experimental Setup}

First, we divided the provided data set into training and test sets (90\% and 10\%).
This results in 57,512 training pairs and 6,391 test pairs for within topics classification as well as 54,943 training and 6,105 test pairs for cross topics taken all from the shared task labeled training data.
Then we used \textsc{BERT}~\cite{DBLP:conf/naacl/DevlinCLT19}\footnote{\url{https://github.com/huggingface/transformers}} in our implementation for training to classify arguments of same stance.
We used both provided models \textit{base} and \textit{large} always with three epochs for fine-tuning. All models use lower-cased token sequences and vocabulary.
It should be noted here that \textsc{BERT} is limited to a fixed size of tokens, with the maximum being 512 tokens for the pre-trained models.
Longer input sequences are truncated to the maximum sequence length.
This truncation can lead to loss of information which we evaluate in hypotheses 3 and 4.
%To avoid this threat, we have always considered only the first 32, 64, 128, 256, and 512 tokens of the arguments under the assumption that the stance of an argument is usually in the beginning.

\subsection{Results and Discussion}

\begin{figure}[htb]
    \centering
    \includegraphics[width=\columnwidth]{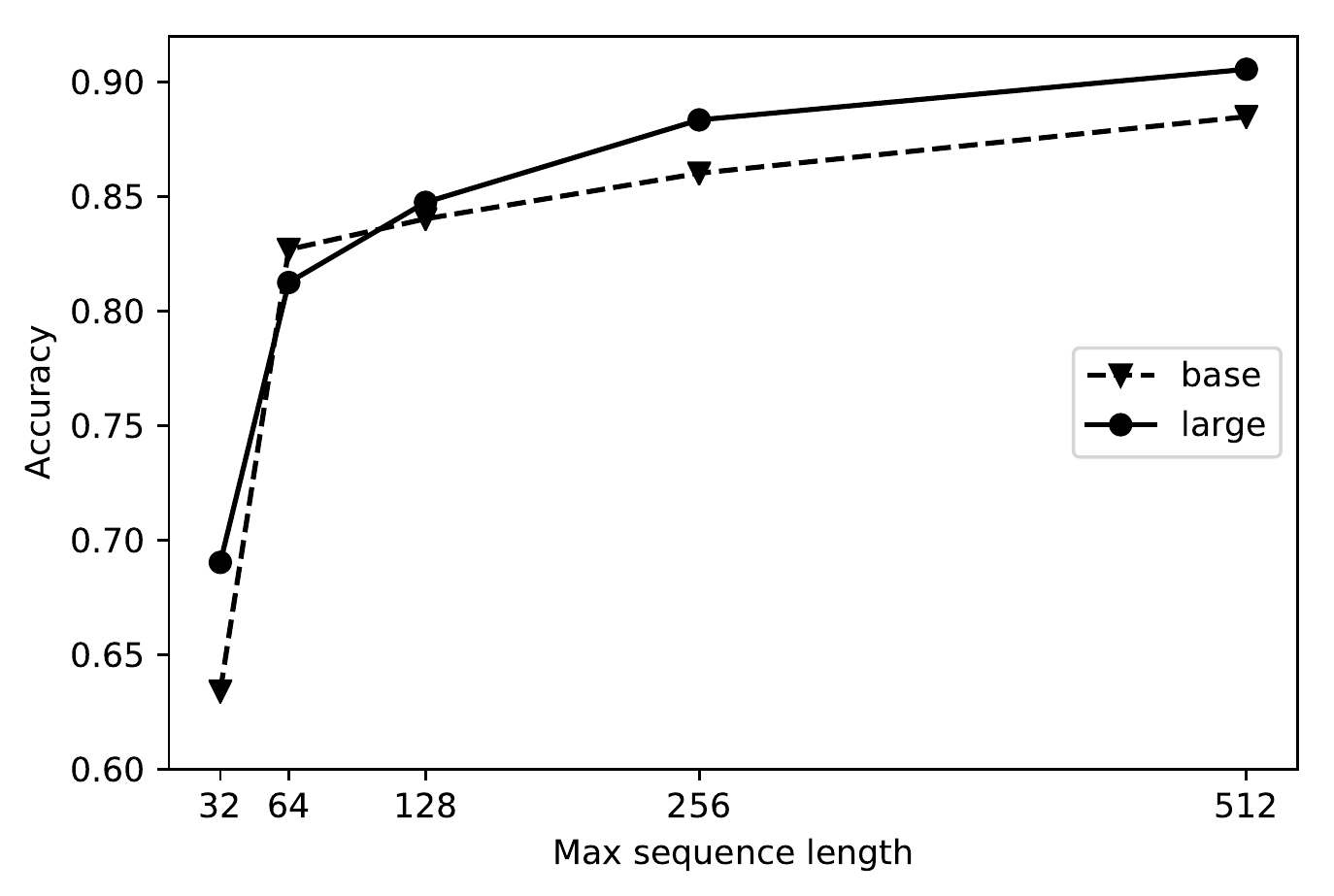}
    \caption{Accuracy of \textit{base} and \textit{large} model \textbf{within} topics for varying maximum sequence length}
    \label{fig:ssc-model-context-stats-within-topics}
\end{figure}

\begin{figure}[htb]
    \centering
    \includegraphics[width=\columnwidth]{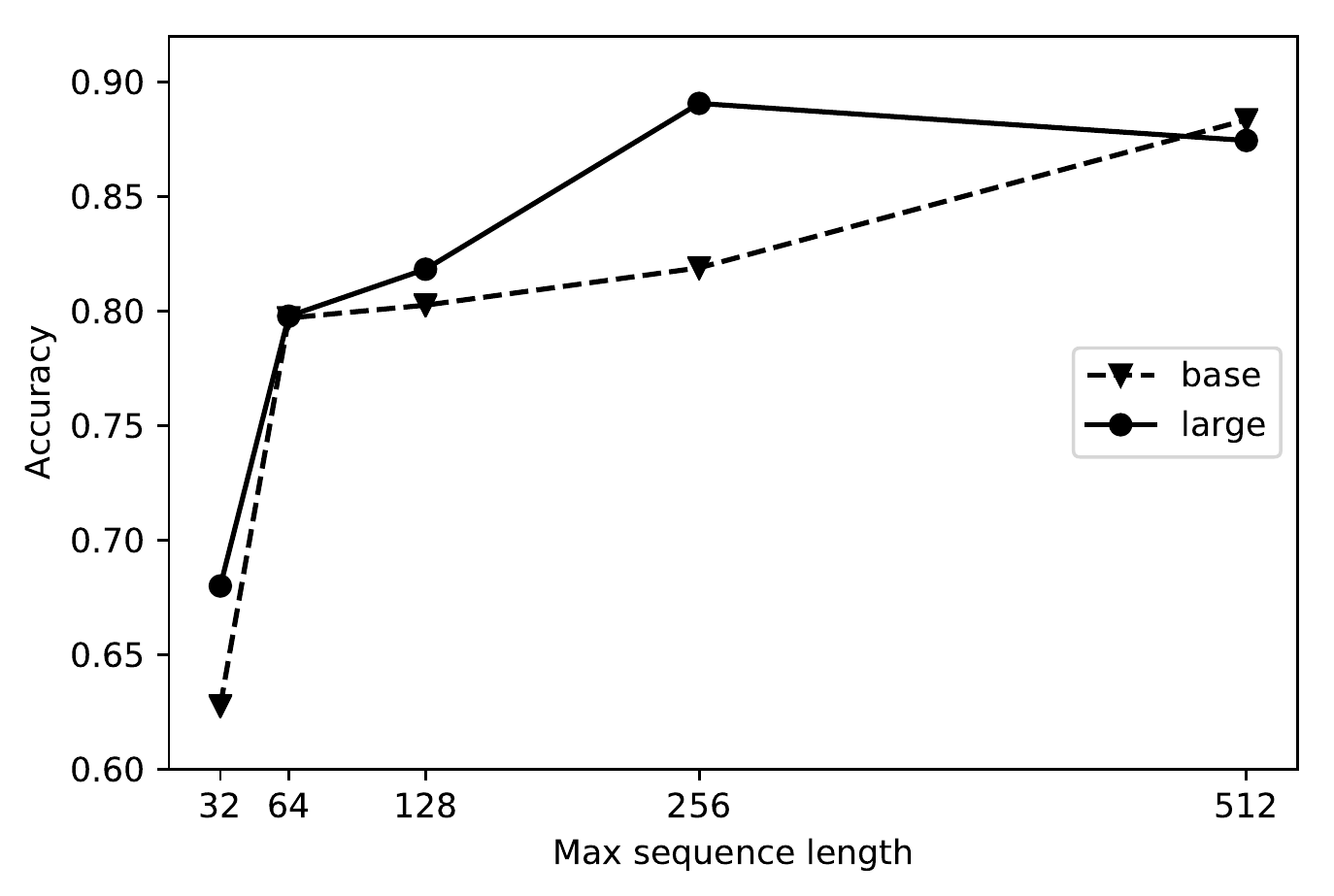}
    \caption{Accuracy of \textit{base} and \textit{large} model \textbf{across} topics for varying maximum sequence length}
    \label{fig:ssc-model-context-stats-cross-topics}
\end{figure}

Figure~\ref{fig:ssc-model-context-stats-within-topics} shows the results within the same topic with varying maximum sequence length.
Figure~\ref{fig:ssc-model-context-stats-cross-topics} shows the results between topics.
Argument pairs whose length exceeds the maximum sequence lengths are uniformly truncated.
In within topic evaluation \textit{large} yields higher accuracy than \textit{base} in four of five cases. In the cross topic evaluation the result is similar.
Therefore hypothesis~2 can be accepted.
% The \textit{large} model has more parameters and seems to be able to make better use of the training data.
Nevertheless the smaller \textit{base} model is quite close to \textit{large}.

The SVM baseline, supplied by the shared task organizers, achieves 54\% accuracy in within topic and 58\% cross topic. The \textit{base} model improves upon this already with the smallest sequence length of 32 tokens. Thus hypothesis~1 can be accepted. This result is possibly due to a Transformer having a larger model capacity and employing better suited representations for natural language text compared to an SVM.

\begin{figure}[htb]
    \centering
    \includegraphics[width=\columnwidth]{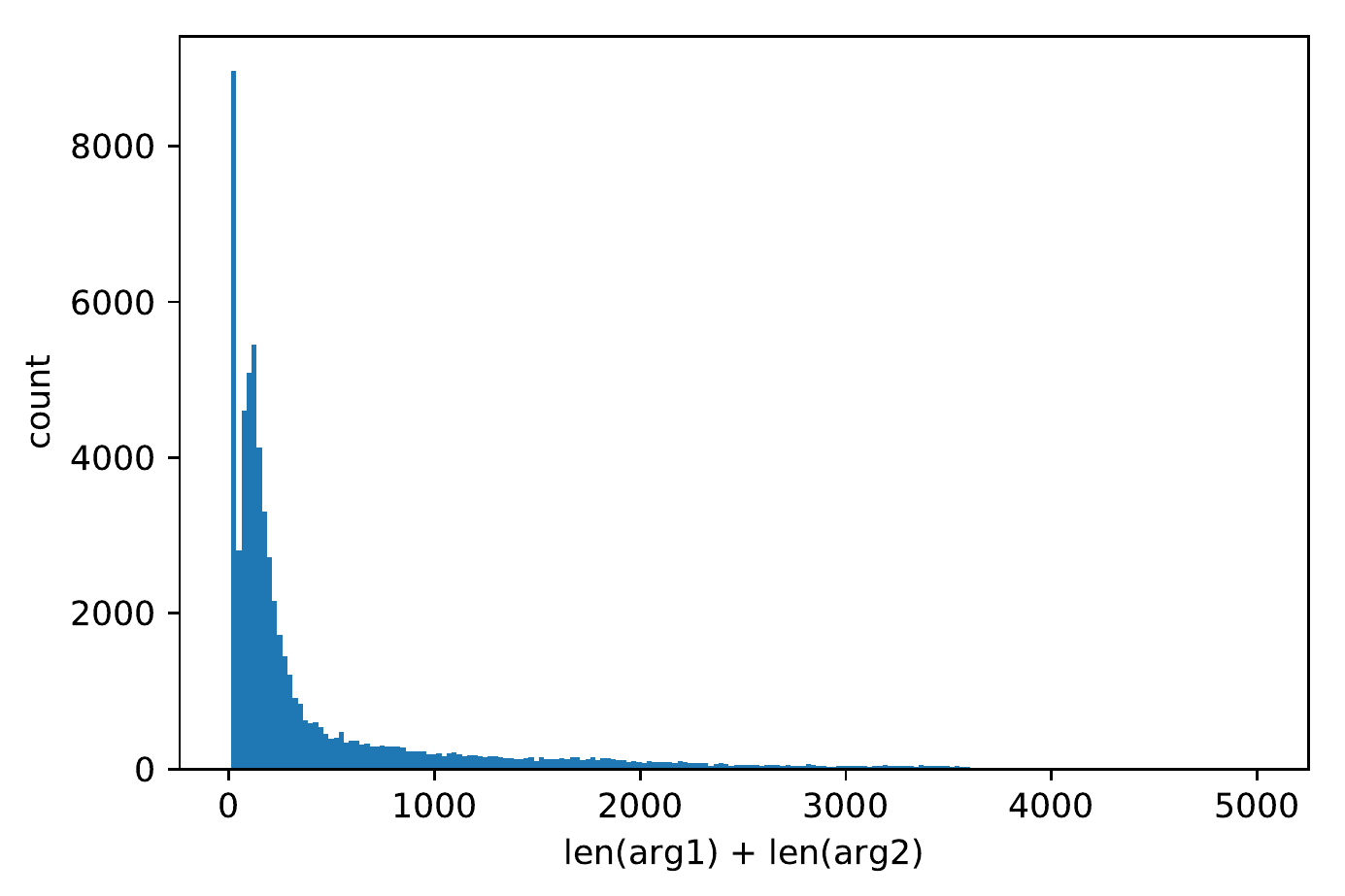}
    \caption{Distribution of argument pair lengths}
    \label{fig:arg-lengths}
\end{figure}

\begin{figure}[htb]
     \centering
     \includegraphics[width=\columnwidth]{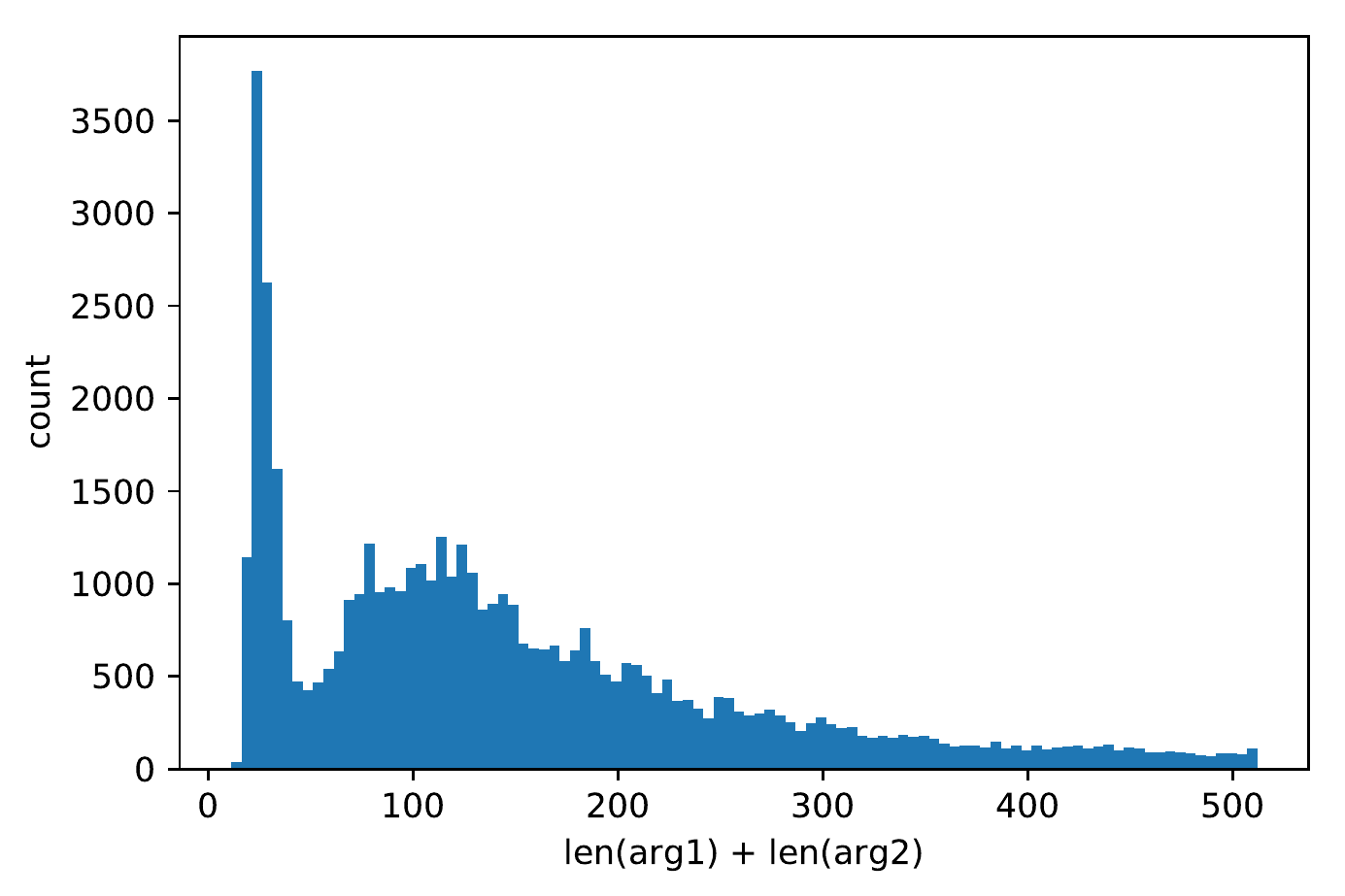}
     \caption{Distribution of token lengths within a maximum of 512 tokens.}
     \label{fig:distributionOfTokenLengthsWithin512Tokens}
 \end{figure}

Next, we take a look at argumentative input of varying maximum sequence length.
We can see from Figure~\ref{fig:ssc-model-context-stats-within-topics} and Figure~\ref{fig:ssc-model-context-stats-cross-topics} that the classification benefits from more contextual information.
Hypothesis 3 can therefore be accepted.
One question is why a model using 64 tokens already performs quite well.
Figure~\ref{fig:arg-lengths} shows the distribution of the summed argument lengths. We can observe here that the majority (76\%) consists of less than 512 tokens.
As we can infer from Figure~\ref{fig:distributionOfTokenLengthsWithin512Tokens}, the distribution of the lengths is usually even considerably below 512 tokens.
This explains why models with rather short contextual information perform quite well already.

Since the input sequences are truncated the Transformer model also trains with incomplete, partial natural language sentences.
In order to see whether a model can better learn from full sentences we filter out all partial sequences which are longer than 512 tokens, reducing the available training/testing data (no trunc train/test).
As can be seen in Table~\ref{tab:resultsWithinTopics} and Table~\ref{tab:resultsAcrossTopics} the Transformer is able to learn from partial sentences.
Therefore hypothesis 4 needs to be rejected.
The highest results are achieved when testing is also done on untruncated full sentences. This result is an indicator of what could be achieved with Transformers of larger or variable maximum sequence length such as explored by \cite{DBLP:conf/acl/DaiYYCLS19}.

\begin{table*}[htb]
\caption{Results of truncated/non-truncated training/testing within topics.}
\label{tab:resultsWithinTopics}
\centering
%\resizebox{\textwidth}{!}{
\begin{tabular}{|c|c|l|l|l|l|l|}
\hline
\textbf{No Trunc} & \textbf{No Trunc} & \textbf{Model} & \textbf{\# Train} & \textbf{\# Test} & \textbf{Acc} & \textbf{F1} \\
\textbf{Train} & \textbf{Test} & & & & & \\
\hline
\hline
   &   & bert-base  & 57,512 & 6,391 & 0.8848 & 0.8911 \\
   &   & bert-large & 57,512 & 6,391 & 0.9055 & 0.9104 \\
 \fatterdot &   & bert-base  & 43,678 & 6,391 & 0.8603 & 0.8668 \\
 \fatterdot &   & bert-large & 43,678 & 6,391 & 0.8830 & 0.8896 \\
 \fatterdot & \fatterdot & bert-base  & 43,678 & 4,932 & 0.9064 & 0.9137 \\
 \fatterdot & \fatterdot & bert-large & 43,678 & 4,932 & \textbf{0.9471} & \textbf{0.9527} \\
\hline
\end{tabular}
%}
\end{table*}

\begin{table*}[htp]
\caption{Results of truncated/non-truncated training/testing across topics.}
\label{tab:resultsAcrossTopics}
\centering
%\resizebox{\textwidth}{!}{
\begin{tabular}{|c|c|l|l|l|l|l|}
\hline
\textbf{No Trunc} & \textbf{No Trunc} & \textbf{Model} & \textbf{\# Train} & \textbf{\# Test} & \textbf{Acc} & \textbf{F1} \\
\textbf{Train} & \textbf{Test} & & & & & \\
\hline
\hline
   &   & bert-base  & 54,943 & 6,105 & 0.8834 & 0.8848 \\
   &   & bert-large & 54,943 & 6,105 & 0.8744 & 0.8757 \\
 \fatterdot &   & bert-base  & 40,763 & 6,105 & 0.8139 & 0.7971 \\
 \fatterdot &   & bert-large & 40,763 & 6,105 & 0.8622 & 0.8667 \\
 \fatterdot & \fatterdot & bert-base  & 40,763 & 4,515 & 0.8997 & 0.9026 \\
 \fatterdot & \fatterdot & bert-large & 40,763 & 4,515 & \textbf{0.9271} & \textbf{0.9325} \\
\hline
\end{tabular}
%}
\end{table*}

\section{Conclusion}
\label{conclusion}
In this paper we have contributed to the $\gg$\textsc{same side (stance) classification}$\ll$ task and proposed a method which uses a fine-tuned \textsc{BERT} model to determine whether two given arguments have the same stance.
The baseline of the organizers was outperformed with our method. In our evaluation the large model performs better than the base model. Our results also show that longer input sentences are classified better than shorter ones, and that classifying whole sentences does not perform better than classifying partial sentences.
According to the organizers' leaderboard~\cite{SameSideWebisLeaderboard}\footnote{Ranking on the 16th August 2019.} our approach performed best across topics with precision and recall values of 0.72 and an accuracy of 0.73.
For within-topics we achieved the best performance as well as the ASV team from Leipzig University with an accuracy of 0.77. However, for this task we had a higher precision (0.85 vs. 0.79) but a lower recall (0.66 vs. 0.73).

\section*{Acknowledgements}
This work has been funded by the Deutsche Forschungsgemeinschaft (DFG) within the project ReCAP, Grant Number 375342983 - 2018-2020, as part of the Priority Program ``Robust Argumentation Machines (RATIO)'' (SPP-1999).

\bibliographystyle{acl_natbib}
\bibliography{main}

\end{document}